\documentclass[letterpaper, 10 pt, conference]{ieeeconf}  

\IEEEoverridecommandlockouts                              

\usepackage{amsmath} 
\usepackage{amssymb}  
\usepackage{algorithm}
\usepackage{subcaption}
\usepackage{adjustbox}
\usepackage{booktabs, multirow}
\usepackage{xcolor}
\usepackage[noend]{algpseudocode}
\usepackage[utf8]{inputenc}
\usepackage{graphicx}
\usepackage{tikz}
\usetikzlibrary{backgrounds}

\title{\LARGE \bf
EXOT: Exit-aware Object Tracker for Safe Robotic Manipulation of Moving Object
}

\author{Hyunseo Kim$^{1}$, Hye Jung Yoon$^{2}$, Minji Kim$^{3}$, Dong-Sig Han$^{3}$, and Byoung-Tak Zhang$^{4}$ 
\thanks{*This work was supported by Institute of Information \& communications Technology Planning \& Evaluation (IITP) grant funded by the Korea government (MSIT) [NO.2021-0-01343, Artificial Intelligence Graduate School Program (Seoul National University)/25\%, 2021-0-02068-AIHub/25\%, 2022-0-00951-LBA/25\%, 2022-0-00953-PICA/25\%)]
}
\thanks{$^{1}$Interdisciplinary Program in Neuroscience, Seoul National University, Seoul, Korea}%
\thanks{$^{2}$Interdisciplinary Program in Artificial Intelligence, Seoul National University}%
\thanks{$^{3}$Dept. of Computer Science and Engineering, Seoul National University}%
\thanks{$^{4}$AIIS, Seoul National University}%
\thanks{{\tt\small \{hskim, hjyoon, mjkim, dshan, btzhang\}@bi.snu.ac.kr}}
}

\begin{document}

\maketitle
\thispagestyle{empty}
\pagestyle{empty}

\begin{abstract}

Current robotic hand manipulation narrowly operates with objects in predictable positions in limited environments. 
Thus, when the location of the target object deviates severely from the expected location, a robot sometimes responds in an unexpected way, especially when it operates with a human. 
For safe robot operation, we propose the EXit-aware Object Tracker (EXOT) on a robot hand camera that recognizes an object's absence during manipulation.
The robot decides whether to proceed by examining the tracker's bounding box output containing the target object. 
We adopt an out-of-distribution classifier for more accurate object recognition since trackers can mistrack a background as a target object. 
To the best of our knowledge, our method is the first approach of applying an out-of-distribution classification technique to a tracker output. 
We evaluate our method on the first-person video benchmark dataset, TREK-150, and on the custom dataset, RMOT-223, that we collect from the UR5e robot. 
Then we test our tracker on the UR5e robot in real-time with a conveyor-belt sushi task, to examine the tracker's ability to track target dishes and to determine the \emph{exit} status. 
Our tracker shows 38\% higher \emph{exit-aware} performance than a baseline method.
The dataset and the code will be released at https://github.com/hskAlena/EXOT.

\end{abstract}

\section{INTRODUCTION}

\begin{figure}
\centering
\includegraphics[width=0.45\textwidth]{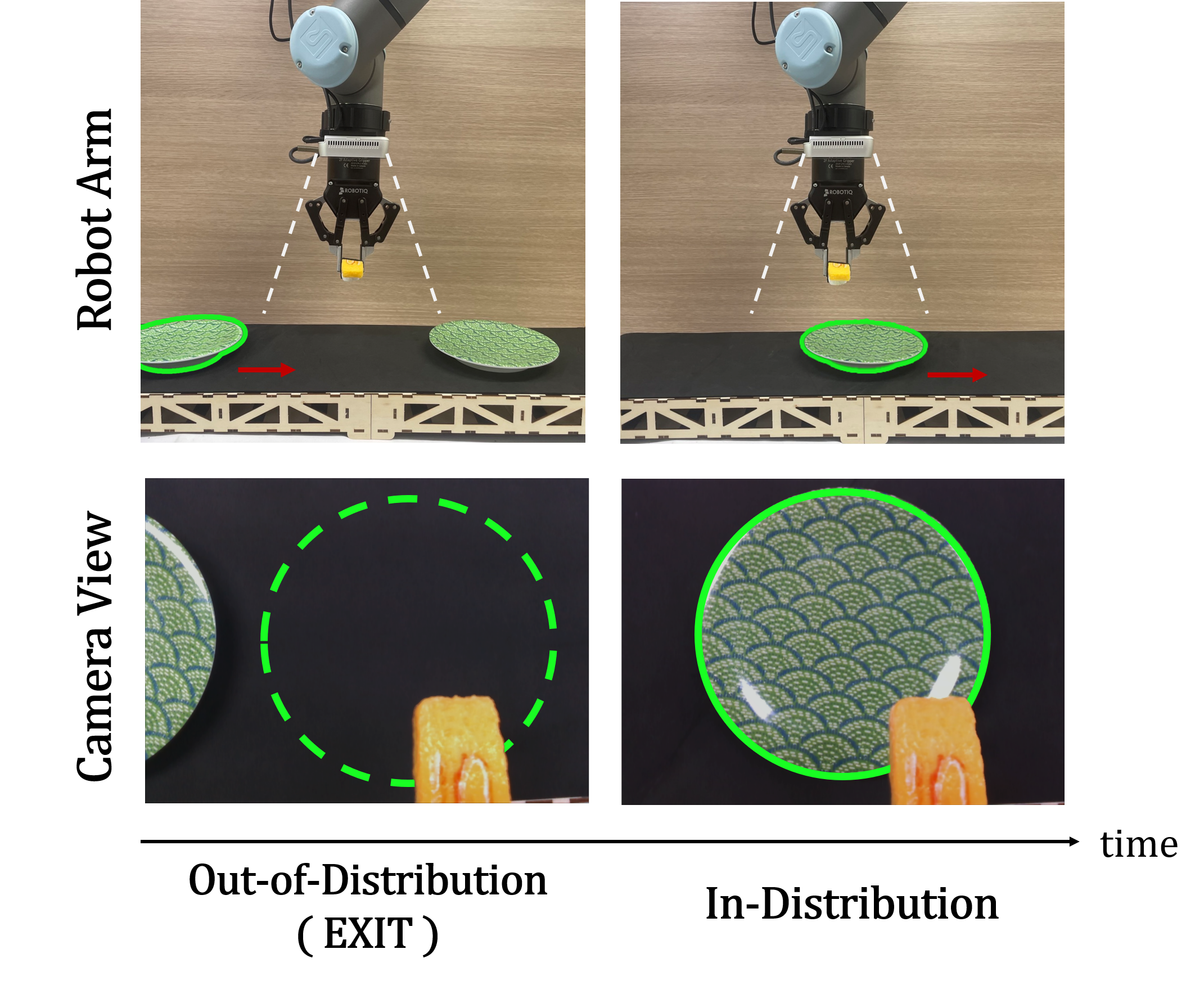}
\caption{\small An overview of the problem setting. A robot with a hand camera tracks down the target object (highlighted with green) on a moving conveyor-belt. Target objects are not evenly positioned on the conveyor-belt, so empty spots come up frequently. 
The robot recognizes an abnormal situation (exit) and does not try placing when the target object is not visible on the camera. When the target object is visible on the hand cam, the robot places the object it was holding. A camera frame with the target object is in-distribution (ID) and the frame without the target object is out-of-distribution (OOD).}
\label{task_overview}
\vspace{-5mm}
\end{figure}

Visual processing of a target object is an essential part of robotic manipulation~\cite{tremblay2018corl:dope,gou2021rgb, danielczuk2020exploratory, camacho2021reward, driess2021learning}. 
In previous works, the position of the target object is once calculated based on a camera view, then the robot arm moves to the position and manipulates the object~\cite{robotnavi, rad2017bb8, xiang2017posecnn, tekin2018real}. Generally, the robot arm moves blindly after calculating the position. 
Therefore, it is difficult to react to target movement when the robot is not in a laboratory environment. 
In this paper, we tackle the problem described in Fig.~\ref{task_overview}. A robot works in a conveyor-belt sushi store and helps a chef by placing sushi on a target dish. Target dishes are unevenly placed on a conveyor-belt, and the robot needs to recognize the absence of the target dish in order to complete the task safely. The sushi store is busy and packed with customers, so the robot can only rely on a hand camera during the process, not having enough space for a ceiling camera that generates a global coordinate. 
 
A problem of tracking down moving objects on a hand camera is similar to a first-person visual object tracking problem~\cite{dunnhofer2021first}.
The visual object tracking~\cite{staple,endToend,Fan_2019_CVPR, kristan2020eighth, huang2019got, kristan2021ninth} domain has been extensively studied and shows outstanding performance in multiple video tasks~\cite{mixformer, stark, lin2021swintrack}. 
Specifically, object tracking in first-person videos~\cite{Damen2018EPICKITCHENS, wang2020symbiotic, liu2020forecasting} has been tackled with long-term object trackers, which resume tracking when a once-invisible target object re-enters a frame. 
Since raw videos from a robot hand camera share similarities with first-person videos, long-term object trackers are solutions to the problem of tracking down moving objects on a robot hand camera. 
We propose a new long-term object tracker, fused with an out-of-distribution (OOD) classifier that enables safe robotic manipulations.

\begin{figure*}
\centering 
{\includegraphics[height=0.35\textwidth]{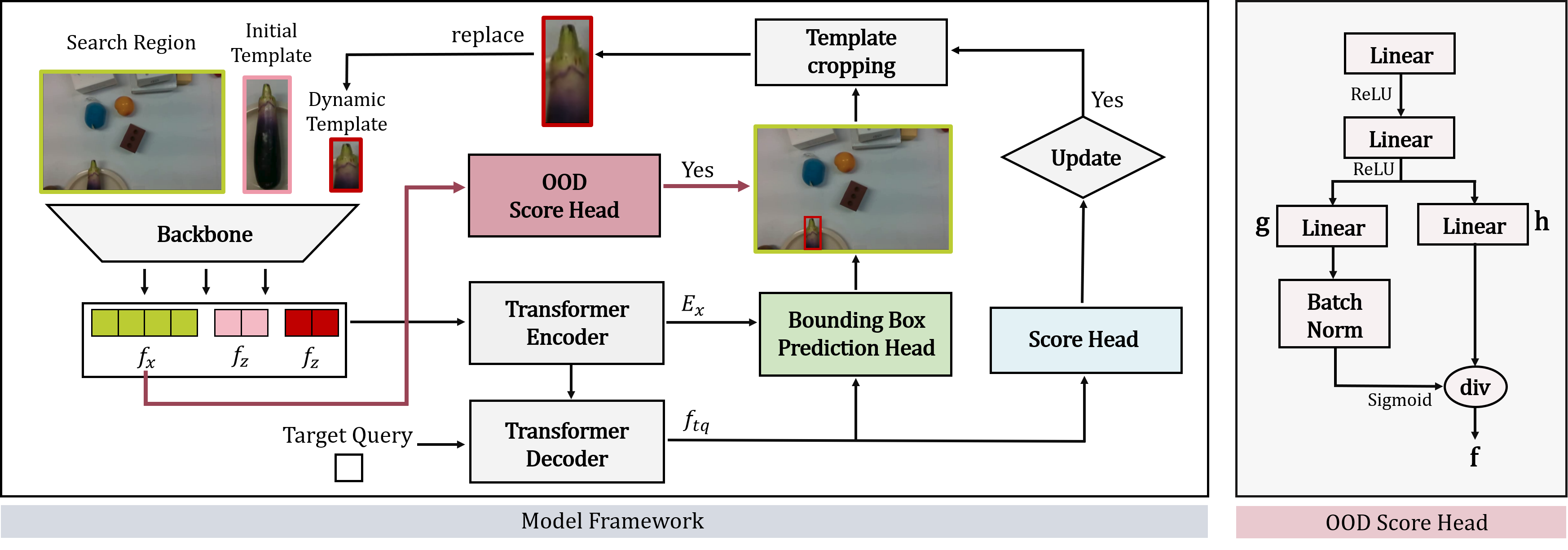}}
 \caption{\small An illustration of the proposed tracker, EXOT. At the start of training, a search image and an initial template are processed with the backbone network and concatenated. Then those are fed into the Transformer Encoder and Decoder sequentially. 
 Three head networks (OOD score head, bounding box (bbox) prediction head, and score head) process inputs described in the diagram and output a classification confidence score, bbox coordinates, and a template update score, respectively. Using the information, a dynamic template is cropped from the search image and used as an input in the next step. 
 The inside networks of OOD score head are described on the right. Further details are explained in Eq.~\ref{eq:logit_ood}.
 }

\label{fig:EXOT}
\vspace{-5mm}
\end{figure*}

Classification in object tracking has mainly been studied in multi-object tracking~\cite{sun2019deep, luo2021multiple,emami2018machine} rather than in single-object tracking. 
Classifiers merely work separately with trackers and do not provide any help related to tracking performance. However, classifiers can work in synergy with single object trackers by providing choices to the trackers to have confidence in the tracking results. 
In this work, we attach an OOD classifier to a single-object tracker to recognize the absence of a target object. The OOD classifier outputs low confidence when an input is an image without the target object.

First, we set up our tracker network based on STARK~\cite{stark}, which is based on the Transformer network~\cite{attention} and dynamically updates a target object template. 
The network works as a long-term tracker by updating and integrating the target information. 
After accurately localizing the target bounding box (bbox), we additionally train an OOD classifier on the tracking result. 
Inspired by the Generalized ODIN~\cite{generalized_ODIN}, we build a network that reflects an input domain distribution and a joint probability between class labels and the domain. 
Consequently, our classifier sorts inputs into in-domain and out-of-domain, which is useful for recognizing the wrongly tracked backgrounds. 

In our experiments, we find that our method successfully recognizes a target object's absence in the benchmark dataset TREK-150~\cite{TREK150} and in the custom dataset, named RMOT-223 dataset, collected with the UR5e robot. Our tracker outperforms the base tracker STARK in several criteria, and we demonstrate our modelling by conducting ablation studies. Moreover, we set up a conveyor-belt sushi setting to show our tracker's outstanding performance of recognizing a target plate's absence.

Our main contributions are summarized as follows:
\begin{itemize}
    \item A robot with our tracker successfully and safely picks and places on a moving target safely even when it operates without a global coordinate. 
    \item We point out that the absence of target objects during tracking outputs unreliable tracking results and propose a way to enhance the performance of a tracker by combining an OOD classifier with a tracker. 
    \item We propose the new robot object dataset, RMOT-223, collected by both automatic robot operation and human tele-operation using the UR5e robot. RMOT-223 can also work as a first-person view tracking dataset with diverse collecting methods.
\end{itemize}


\section{BACKGROUNDS}
\subsection{Object Tracking}
Visual object tracking, which infers the changing states of a target object as a video proceeds, is divided into two major parts: single object tracking~\cite{Fan_2019_CVPR, TREK150} and multi-object tracking~\cite{milan2016mot16, bai2021gmot}. 
In robotics, a single object tracker is more useful because a robot grasps one target object at a time. In addition, it has the ability to track unseen objects without learning, which makes robot operation more stable as object detection results may be unstable due to external factors (i.e. lights).
However, the use of a single object tracker on a robot hand camera should be carefully considered, because a target object may more easily be out of sight of the camera as the robot hand approaches it. 
The tracker must be \emph{exit-aware} to deal with this problem. In other words, the tracker should recognize the absence of the target object and send a signal to the robot to take an appropriate action. 

The tracker should also have the long-term tracker's characteristic, which is to memorize the previous target appearance and decide whether to update the target image in every few time steps. Long-term trackers may indirectly know the absence of the target when they decide whether to update the target image, but they cannot explicitly signal the absence to the robot. Also, as the target template update module is not trained for recognizing the target absence, it is unreliable to use it as the absence notifier. Therefore, an additional module that decides the target absence is needed, which we implemented using an OOD classifier.

\begin{table}
\caption{\small The statistics of exit frequency and object diversity in datasets. (EVR: Exit video ratio, AEL: Average length of exit frames in one video, AVL: Average length of video frames, MIEL: Minimum length of exit frames, MAEL: Maximum length of exit frames)}
\label{custom_analysis}
\begin{center}
\begin{tabular}{c|c|c|c|c|c|c} 
 \hline
  & EVR & AEL & AVL & MIEL & MAEL & \#obj  \\ [0.5ex] 
 \hline\hline
  TREK-150 & 0.26 & 56.46 & 722.0 & 1 & 336 & 34 \\  
 \hline
  RMOT-auto & 0.112 & 51.24 & 303.08 & 19 & 191 & 19 \\ 
 \hline
  RMOT-human & 0.026 & 161.67 & 711.5 & 20 & 565 & 19 \\ 
 \hline
\end{tabular}
\end{center}
\vspace{-5mm}
\end{table}

In addition, datasets with an abundant number of exit instances are used to train a model for learning \emph{exit-awareness}. TREK-150~\cite{TREK150} is a first-person view object tracking dataset that originated from the EPIC-KITCHEN dataset~\cite{Damen2018EPICKITCHENS}. The dataset has plentiful object exit circumstances since videos are recorded from the head mounted camera of people who spend time in the kitchen. 
Also, the camera viewpoint is first-person view, the same as the robot hand camera. In this paper, we used TREK-150, and we collected the RMOT-223 dataset, which follows the annotation rule of TREK-150, to complement the lack of other first-person view object tracking datasets. 

\subsection{Out-of-distribution Detection}
In real-world situations, we often encounter unseen objects, the objects that are occasionally unknown. A well-trained classifier should detect such instances as anomalous and make a conservative decision. In particular, when \emph{unknown} is not listed among the available labels, the well-trained classifier classifies the unknown objects with low confidence, indicating that the samples are anomalous. 
However, naive classifiers classify them with high confidence under the wrong label~\cite{oodsurvey}. To classify unknown objects with low confidence and vice versa, the sample distribution of known objects, which is a training set, should be clarified in the classifier. 
In other words, the unknown objects are recognized as \emph{unknown} because they are out of the training set distribution in the well-trained classifier's viewpoint. Therefore, they are called out-of-distribution (OOD) samples, and the task of labeling them as \emph{unknown} is OOD detection. 

OOD samples are called in several ways: anomalies, novelties, and outliers~\cite{oodsurvey}. Since OOD is defined in diverse ways, there are a lot of ways to deal with it. In this paper, we focus on the classification-based technique. We defined OOD sample as an image without the target object and in-distribution (ID) as an image with the target object, as shown in Fig.~\ref{task_overview}.

\begin{equation}
\label{eq:domain_ood}
    p(y|d_{in}, x) = \frac{p(y, d_{in} | x)}{p(d_{in} |x)} 
\end{equation}
Generalized ODIN~\cite{generalized_ODIN} is a method that trains the classifier to correctly define the sample distribution. Conventional classifiers learn to classify the objects within the given labels. They only learn the class posterior probability $p(y | d_{in}, x)$, where $y$ is the class label and $d_{in}$ means that the input $x$ is within the training set domain. However, in order to know the training distribution clearly and thus identify the OOD samples, the classifier should learn the domain probability $p(d_{in} | x)$. Therefore, using the Eq.~\ref{eq:domain_ood}, the Generalized ODIN divides the classifier network into two, one predicting  $p(y, d_{in} |x)$ and the other predicting $p(d_{in} | x)$.

\begin{figure}[t]
    \centering
    \begin{subfigure}[b]{0.48\textwidth}
        \centering
        \includegraphics[width=\textwidth]{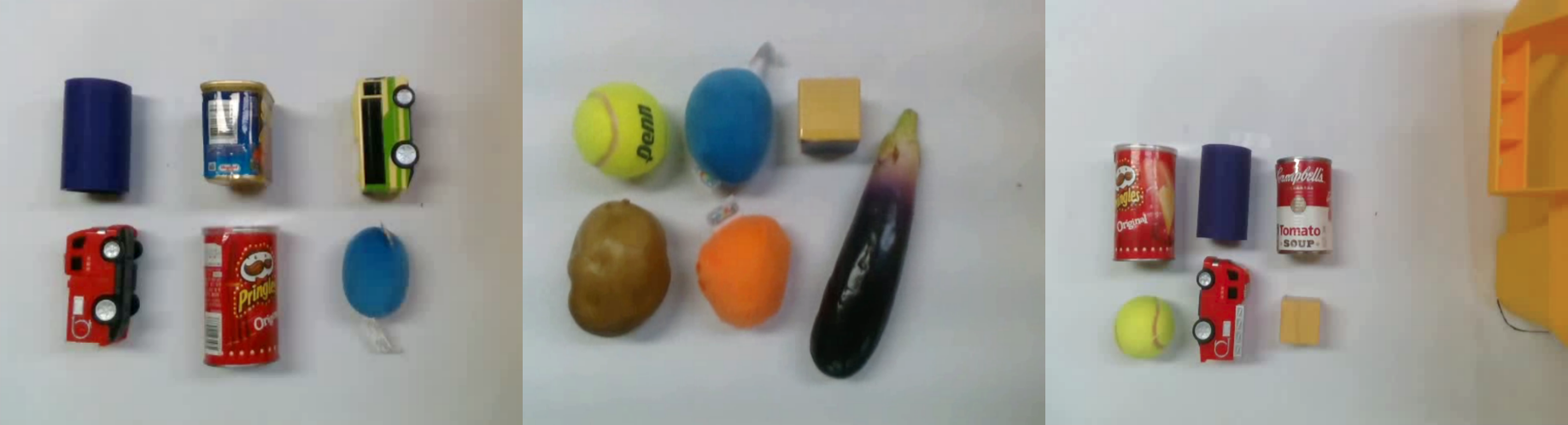}
        \caption[]%
        {{\small RMOT-223 dataset}}
    \vspace{-5mm}
    \end{subfigure}
    \hfill
    \vskip\baselineskip
    \begin{subfigure}[b]{0.48\textwidth}  
        \centering 
        \includegraphics[width=\textwidth]{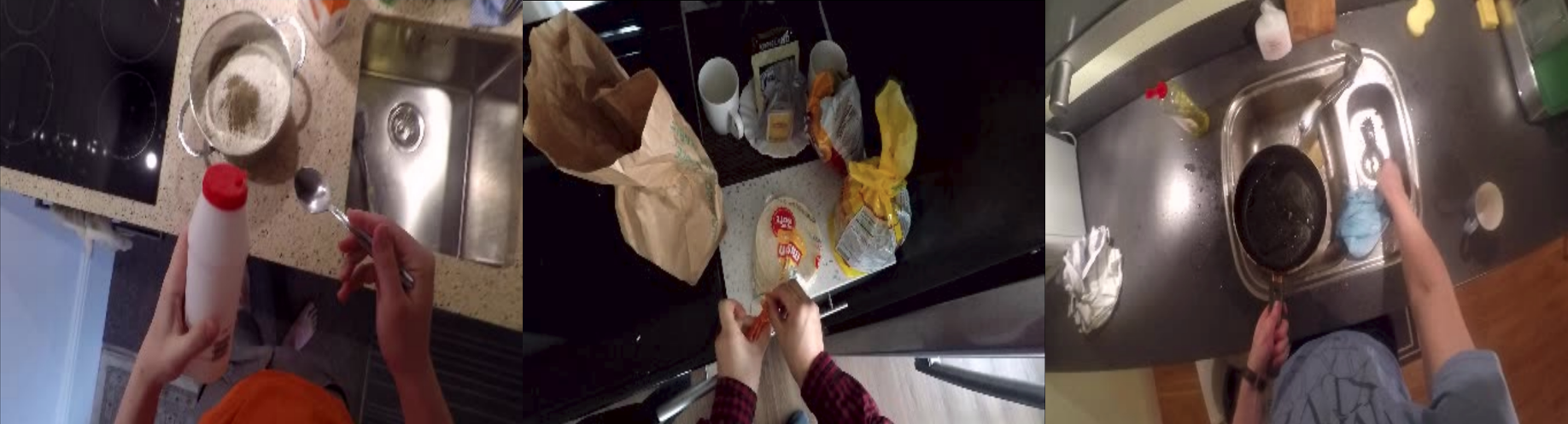}
        \caption[]%
        {{\small TREK-150 dataset}}    
    \end{subfigure}
    \caption[]
    { \small Dataset examples. (a) shows how objects are positioned in the pick-and-place task. In average, 6 objects are shown in one scene and each video is recorded from the start of robot moving and until the robot picks up one object. 
    (b) is examples from TREK-150 dataset. }
    \label{dataset}
\vspace{-5mm}
\end{figure}

\section{Exit-aware Object Tracking}
In this section, we propose an exit-aware object tracker, called EXOT. As described in Fig.~\ref{fig:EXOT}, EXOT is divided into 3 parts: backbone, Transformer, and heads. 
Inherited from our base tracker STARK, the backbone network is constructed with a vanilla ResNet, and the Transformer network has the structure similar to that of DETR~\cite{detr}, a fancy object detection network. 
The role of the three heads will be explained in the following subsections.

\subsection{Bounding box prediction}
\label{sec:bbox}

First, a template image $z$ and a search region of a current frame $x$ are converted into backbone features $f_z, f_x$, respectively. Then, these features are concatenated and fed into the Encoder-Decoder Transformer, which outputs the Encoder output of the search region $E_x$, and a target query feature $f_{tq}$. 
For the bounding box (bbox) prediction, a similarity score $s$ between $f_{tq}$ and $E_x$ is computed, indicating which part of the search image has similarity with the target template. 
After narrowing down the similarity region with convolutional neural networks (CNN), the bbox coordinate is corner-separately computed by multiplying the similarity with the search image coordinate. 
The evaluation is carried out with the ground truth value using L1 loss and generalized intersection of union (GIoU) loss.

\begin{figure}[t]
    \centering
    \begin{subfigure}[b]{0.47\textwidth}   
        \centering 
        \includegraphics[width=\textwidth]{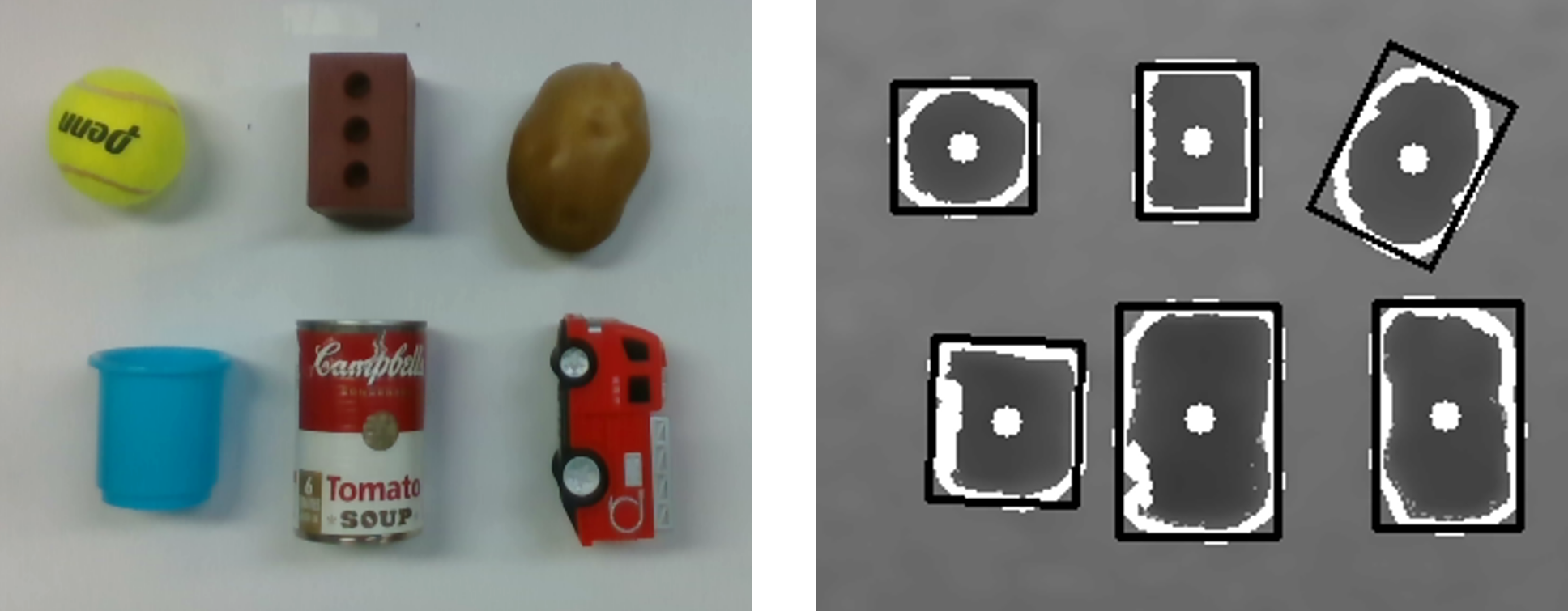}
        \caption[]%
        {{\small Depth-contour image }}    
    \end{subfigure}
    \vspace{-5mm}
    \hfill
   \vskip\baselineskip
    \begin{subfigure}[b]{0.235\textwidth}
        \centering
        \includegraphics[width=\textwidth]{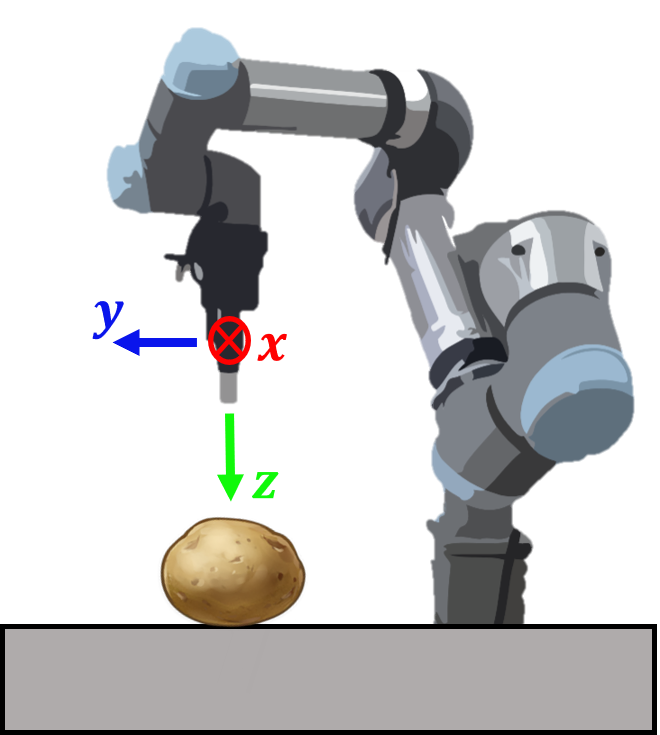}
        \caption[]%
        {{\small }}
        
    \end{subfigure}
    \hfill
    \begin{subfigure}[b]{0.235\textwidth}  
        \centering 
        \includegraphics[width=\textwidth]{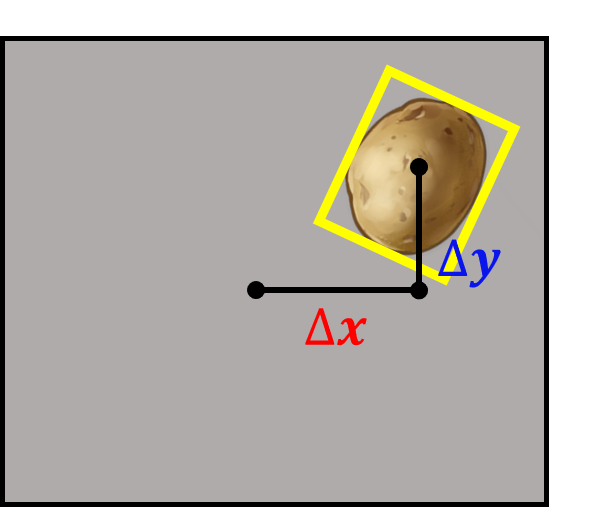}
        \caption[]%
        {{\small }}    
    \end{subfigure}
        
    \caption{ \small An illustration of how our robot gets action signal from its hand camera image. 
    The positive x-y-z axis of robot hand is shown in (b). Then, when a hand camera image is shown like (c), we can get the center pixel point of a target object and the ratio between the pixel width of the camera view and the real world width filmed in the camera view. 
    Using the information, the delta distance the robot should move is deduced to have the target object's center point right under its gripper. Real camera image and its grayscaled image with depth center point are shown in (a).}
    \label{fig:robot_coord}
\end{figure}

\subsection{Template update score prediction}
\label{sec:score}
The template update score is calculated by a simple multi-layer perceptron (MLP), which outputs a regression result of $f_{tq}$. 
If the score is greater than 0.5, a dynamic template is created and used as the template in the next step. 
During training, the score is compared to the ground truth with a binary cross-entropy loss, as each input image is labeled with 0 or 1 to indicate the absence or presence of a target object. 
Since the template update score prediction is trained with a visible label of a target object, it can be used as an exit prediction of the baseline tracker. 

\subsection{OOD score prediction}
\label{sec:ood}
In this section, we present the OOD score prediction head, the main part of our tracker EXOT. 
Inspired by Generalized ODIN, the OOD score head network $f_i (x)$ in Eq.~\ref{eq:logit_ood} is an implementation of the class posterior probability $p(y | d_{in}, x)$ shown in Eq.~\ref{eq:domain_ood}. The OOD score prediction head is a modified MLP network as shown in Fig.~\ref{fig:EXOT}, which outputs the logit $f_i (x)$ for class $i$.
Although EXOT is a single object tracker, the OOD score head has the capability of classifying multiple classes because multiple classes of objects are trained simultaneously as a batch, and training to \emph{classify} within a class degrades the classifier's performance.
Following the design in Eq.~\ref{eq:domain_ood}, instead of mapping the logit $f_i (x)$ with a network, the $h_i (x)$ and $g(x)$ networks each predict the joint class-domain probability $p(y, d_{in} |x)$ and the domain probability $p(d_{in} | x)$.

\begin{equation}
\label{eq:logit_ood}
    f_i (x) = \frac{h_i (x)}{g (x)}
\end{equation}

The logit score in the OOD classifier can work as the OOD score function $S(x)$, and usually the score function is the maximum of $h_i(x)$ or $g(x)$ (Eq.~\ref{scoreood}). 
However, deciding whether a sample is OOD or not on the basis of a single trial can be risky.
Therefore, the perturbation of the inputs is conducted based on the OOD score calculated so far. 
If a sample has a high OOD score, it will receive a higher perturbation, which makes ID samples gather together and OOD samples separate away from ID samples (Eq.~\ref{perturb}). To adjust the perturbation strength, various $\epsilon$ are tried, e.g. $[0.0025, 0.005, 0.01, 0.02, 0.04, 0.08]$. 
After the input is manipulated ($\hat{x}$), the selection of the hyperparameter $\epsilon^*$ (see Eq.~\ref{opteps}) continues, finding the appropriate $\epsilon$ that increases the score of OOD.

\begin{gather}
     S(x) = max_i h_i (x) \operatorname*{or} g(x)  \label{scoreood}\\
    \hat{x} = x - \epsilon \operatorname*{sign}(- \bigtriangledown_x S(x)) \label{perturb} \\
    \epsilon^{*} = \operatorname*{argmax}_\epsilon \sum\limits_{x \in D^{val}_{in}} S(\hat{x}) \label{opteps}
\end{gather}

The perturbation process (Eq.~\ref{scoreood} - \ref{opteps}) takes place after the training of the OOD classifier. The $\epsilon^*$ is selected with validation data and during the test time, the $\epsilon^*$ works as $\epsilon$ in Eq.~\ref{perturb}. At test time, a hyperparameter $\phi$ should be set that works as an OOD threshold to classify test samples by OOD score $S(\hat{x})$. 
To select the threshold $\phi$, the OOD score function $max_i h_i(x)$ must be stabilized and consistent. The score function does not constrain the outputs to be within a range of certain numbers, so the output scores can vary widely.
Also, considering that our input to an OOD classifier is time series data, OOD scores should have consistency. 
Therefore, the moving average of OOD scores under a certain time window is finalized as OOD scores, and the threshold $\phi$ to indicate exit prediction should reflect the change in scores. 

Our OOD network differs from the template update score prediction network, in the sense that we use the feature of the search image $f_x$ as input. We implement some ablation studies regarding the type of an input, detailed in Section~\ref{sec:exp}.

\begin{table*}
  \centering
  \caption{\small Bounding box and exit performance analysis. 
  FPR: False Positive Ratio. AUC: Area Under Curve, the probability that the classifier rank the randomly chosen positive example higher than the randomly chosen negative example. OP75: the point where the threshold set overlaps the 75\% of AUC. 
  $P_{norm}$ stands for the normalized precision, the normalized L1 distance measure between the center of predicted bbox and the center of the groundtruth bbox. 
  Bold numbers indicate the best performance with respect to the metric. 
  }\label{tab:exotmetric}
  \vspace*{3pt}
  \begin{adjustbox}{width=0.9\textwidth,center}  
  \begin{tabular}{c|c|c|c|c|c|c|c|c}
  \toprule
 Dataset & Metric   & EXOTm  & EXOTm-$s$ & EXOT & EXOT-$e$ & EXOT-$s$ & EXOT-$tq$ & STARK\\\midrule
 \multirow{6}{*}[-2pt]{TREK-150} 
 & FPR       & \textbf{0.82} &  0.98 & 0.91 & 0.99 & 0.94 & 0.98 & 0.97\\
 & AUROC    & \textbf{0.41} &  0.35 &  0.17 & 0.10 & 0.25 & 0.14 & 0.03\\ \cmidrule{2-9}
 & AUC (\%)        & 66.58  & 67.39 & 22.93 & 22.85 & 25.71 & 22.53 & \textbf{69.33}\\
 & OP75 (\%)           & 66.31  & 63.82 & 10.41 & 10.65 & 9.71 & 10.31  & \textbf{68.56}\\
 & $P_{norm}$(\%)  & 87.27  & 89.31 & 30.14 & 29.90 & 35.31 & 30.01 & \textbf{90.27}\\ \midrule
\multirow{6}{*}[-2pt]{RMOT-223} 
& FPR              & 0.78  & 0.74 & 1.00 & 0.86 & \textbf{0.71} & 0.96 & 1.00\\
& AUROC            & 0.25  & 0.38 & 0.08 & 0.22 & \textbf{0.45} & 0.22 & 0.00\\ \cmidrule{2-9} 
& AUC (\%)        & \textbf{74.56}  & 72.64 & 73.55 & 70.31 & 72.73 & 73.08 & 71.25\\
& OP75 (\%)           & \textbf{80.76}  & 78.07 & 79.02 & 74.93 & 78.03 & 78.16  & 75.94\\
& $P_{norm}$(\%)  & \textbf{97.85}  & 95.57 & 97.56 & 93.44 & 96.50 & 96.94 & 94.00\\
  \bottomrule
  \end{tabular}
  \end{adjustbox}
\vspace{-4mm}
\end{table*}

\section{EXPERIMENTS}
\label{sec:exp}
In this section, we first introduce the RMOT-223 dataset we collected. The detailed statistical analysis of the RMOT-223 dataset and how we collect RMOT-auto and RMOT-human with the UR5e robot will be explained.
Then, the performance of trackers on bbox prediction and exit prediction is reported on two datasets, TREK-150 and RMOT-223 dataset. 
 For an insightful analysis, we perform additional ablation studies on five variants of our tracker. Finally, we analyze the real robot experiments operated in the conveyor-belt sushi task. By showing the real output image of the trackers, we prove that our tracker can work successfully in real-world situations.

\subsection{RMOT-223 Dataset}

We collect a custom robot-object dataset in the pick-and-place scenario. This dataset is collected to fill the dataset gap since there is no other generic object tracking dataset on first-person view than TREK-150. 
Also, TREK-150 is derived from the part of the EPIC-KITCHEN dataset, so the background of each frame is too complicated to learn the absence of the target object (Fig.~\ref{dataset}(b)). In addition, the camera view changes too quickly, so training the tracker only on this dataset actually hinders the training, as too harsh setting will confuse the tracker.

The RMOT-223 dataset consists of 129 videos of automated pick-and-place activities and 94 videos of tele-operated pick-and-place activities, for a total of 223 videos. We use a UR5e robot to collect the dataset, and teleoperate the robot with a 3D mouse from Connexion and a control program from Radalytica. 
For automatic pick-and-place, we normalize depth images obtained from an Intel RealSense D435i camera which is attached to the robot's hand and compute object contours that indicate the area above the predefined depth threshold. The example of normalized depth and calculated depth contours is shown in Fig.~\ref{fig:robot_coord} (a), the objects with black edges and white center points on are the detected objects that the robot tries to pick and place. The center of the contour is used as a relative action command as shown in the Fig.~\ref{fig:robot_coord}. 
In order to translate the pixel coordinates on the image to the relative coordinates in the viewpoint of the robot hand, the scale and some bias calibration are needed beforehand. 
In terms of human teleoperation, the data collection starts from the same initial position as the automatic operation, and the activities are recorded in the fixed length.

We annotate bboxes in our dataset using MixFormer-1k~\cite{mixformer}, which is a pre-trained long-term object tracking model trained with ImageNet-1k. The format of the bbox is [top-left x, top-left y, width, height]. At the beginning of the video, the annotator specifies the initial position of bbox and MixFormer tracks down the object. When the object is out of sight, the annotator replaces the bbox coordinates inferred by MixFormer with $[-1, -1, -1, -1]$, following the convention suggested by TREK-150. Also, if the inferred bbox from MixFormer is wrong, the annotator corrects the bbox to the new bbox for the tracker to follow. 

The examples of objects and their positions in the custom dataset are shown in Fig.~\ref{dataset} (a). The number of object classes is 34 in TREK-150 and 19 in the custom robot dataset. 
The test set has videos with target objects missing while the training set does not have.
The statistical analysis of the datasets is shown in Table~\ref{custom_analysis}. 
Both datasets have about 20\% of videos that have exit instances, referring to the exit video ratio (EVR) metric in the table.  
The EVR is higher in RMOT-auto because the target objects often leave the camera frame when the robot changes its position to an object-graspable position.  
Considering that the shortest template update frequency of STARK is 10 frames, the average exit length (AEL) is long enough to reduce the tracking performance if the tracker updates the template at the wrong moment.

\subsection{Bounding box prediction performance}
 
We train EXOT in 2 different settings, 1-stage and 2-stage, denoted EXOTm and EXOT, respectively. The variation within a setting is shown after the dash(-), which indicates the difference in the input type of the OOD score head (see Table~\ref{tab:exotmetric}).

According to STARK~\cite{stark}, 2-stage training is recommended because the learning of bbox prediction and the score prediction can interfere with each other, leading to suboptimal solutions~\cite{revisitDetect, revisitRCNN}. Therefore, they proposed the separate training of the bbox prediction and the score predictions (template update scores and OOD scores). The bbox prediction networks are frozen while learning the score predictions.
However, in Table~\ref{tab:exotmetric}, simultaneous training of bbox and score predictions generally improved the bbox and exit prediction performance, and 2-stage models showed significantly lower scores on the TREK-150 dataset compared to the others.
It seems that the classification task worked in synergy with the localization task. Especially when there are objects with similar color to the target, classification will help to concretize the image features.

To analyze which feature of the search image $x$ is the most efficient for the OOD score prediction, we varied the input types of the OOD score head. EXOTm and EXOT input the backbone feature $f_x$ into the OOD score head to define OOD directly from the current search image (Fig.~\ref{fig:EXOT}). 
Since the target object is located in the part of the search image, paying attention to the target object can improve the OOD detection performance. EXOT-$e$ focuses on how the Transformer network conveys the target-related information to the search image, using the Encoder feature $E_x$ as an input. 
EXOT-$s$ and EXOTm-$s$ use the similarity score $s$ as an input, similar to the bbox prediction head, because the target object is usually situated inside the bbox.
Finally, EXOT-$tq$ defines the target query feature $f_{tq}$ as an input, similar to the template update score prediction head, which indirectly learns the absence of the target object.
Although we varied the input type to the OOD score head, the original implementation using the backbone feature $f_x$ is better than others. Thus, we conclude that the original information of the image is more useful for OOD classification, not the image feature mixed with attention weights. 
More details about the experiments will be published online.

\begin{figure}[t]
    \centering
    \begin{subfigure}[b]{0.235\textwidth}
        \centering
        \includegraphics[width=\textwidth]{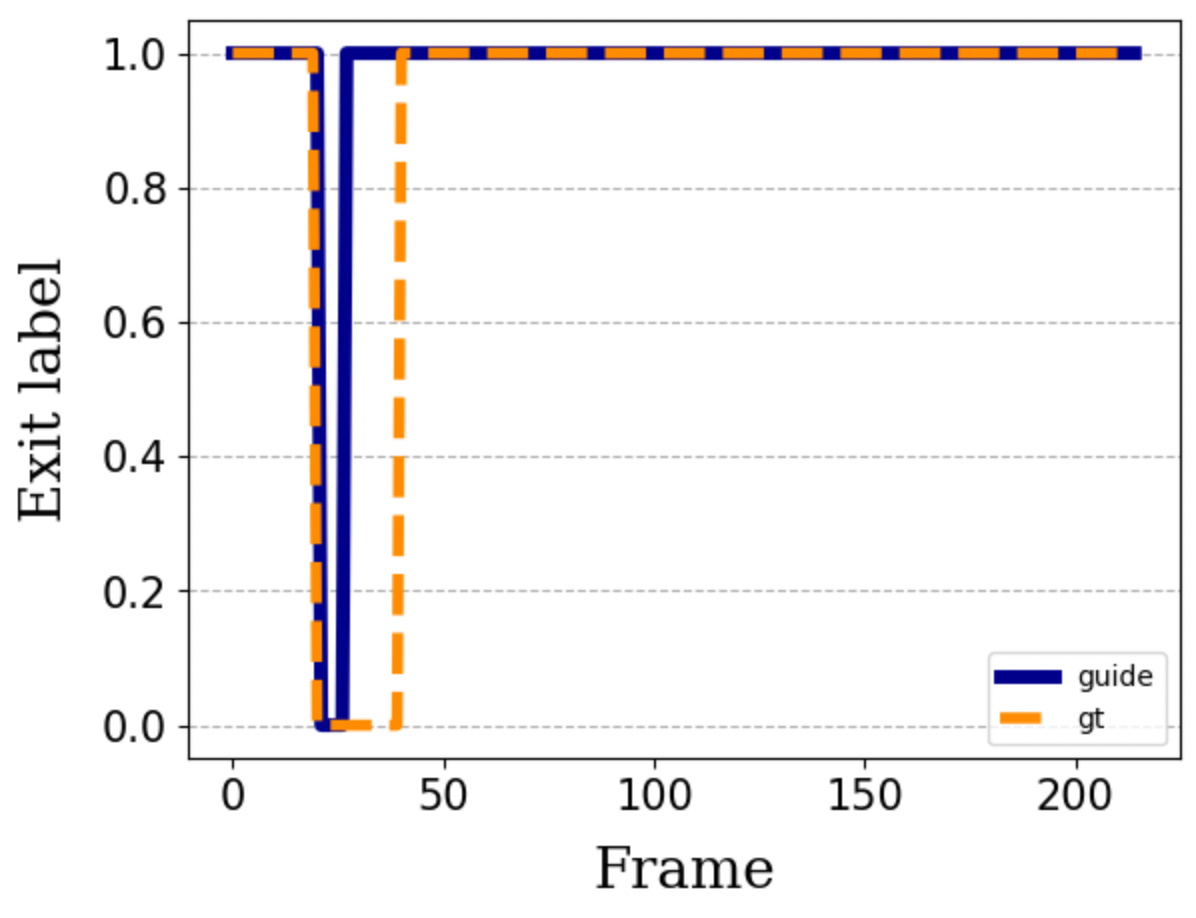}
        \caption[]%
        {{\small STARK prediction }}    
    \end{subfigure}
    \hfill
    \begin{subfigure}[b]{0.235\textwidth}  
        \centering 
        \includegraphics[width=\textwidth]{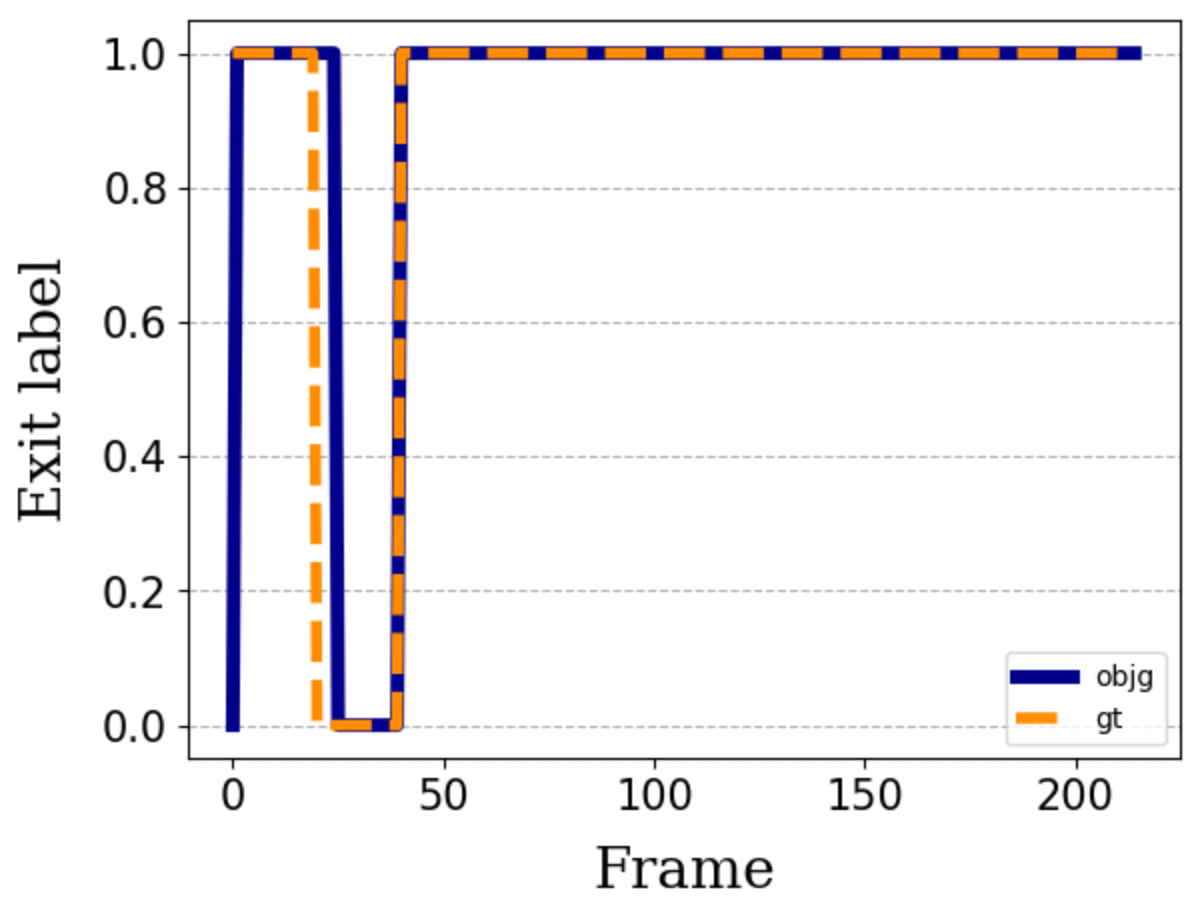}
        \caption[]%
        {{\small EXOT prediction}}    
    \end{subfigure}

    \caption[]%
{\small Qualitative results on exit prediction in RMOT-223 dataset. (a), (b) are exit prediction by STARK and EXOT, respectively. A green line shows exit prediction from the model and a dotted red line shows the groundtruth exit. The groundtruth binary prediction is 1 when the target is visible and 0 when it exits from the frame.} 
    \label{fig:exit_graph}
\vspace{-5mm}
\end{figure}

\subsection{Exit performance on datasets}
In this section, we compared the exit performance of EXOT and STARK qualitatively and quantitatively. 
The exit labels are predicted after calculating the OOD score. The OOD scores are converted to binary exit labels using a threshold $\phi$. Similarly, the template score prediction is converted to the binary exit prediction using 0.5 as the threshold~\cite{stark}. 

See Fig.~\ref{fig:exit_graph}, the exit prediction from EXOT has a graph shape more similar to the ground truth exit graph than the exit prediction from STARK. 
It is important to predict the exit instances conservatively, especially in robotics. The goal of OOD detection is to achieve a higher true negative ratio (TNR) conventionally, aiming to detect all exit instances. However, to emphasize the safety and conservatism, our goal is to lower the false positive ratio (FPR), not allowing exit instances to remain undetected. Two goals are essentially the same, since 1 - TNR = FPR. 
In this sense, we can show in Fig.~\ref{fig:exit_graph} that EXOT performs better in exit prediction, since the error in recognizing exit instances as normal is lower than STARK. 

Also, we use the criteria FPR and AUROC in the quantitative comparison (Table~\ref{tab:exotmetric}). 
AUROC is another name for AUC, but we named it differently to avoid confusion between AUC in bbox prediction and AUROC in exit prediction.
AUC in bbox prediction divides predicted bboxes into positives and negatives, referring to a degree of intersection over union (IoU) with the ground truth bbox. AUROC in exit prediction uses exit labels as negative and non-exit labels as positive.
EXOT performs better than STARK in exit prediction in every case, and we can see that EXOT can robustly detect OOD well in dynamic backgrounds such as the TREK-150 dataset.

\subsection{Safe Robotic Manipulation experiment}
\begin{figure}
\centering
\includegraphics[width=0.5\textwidth]{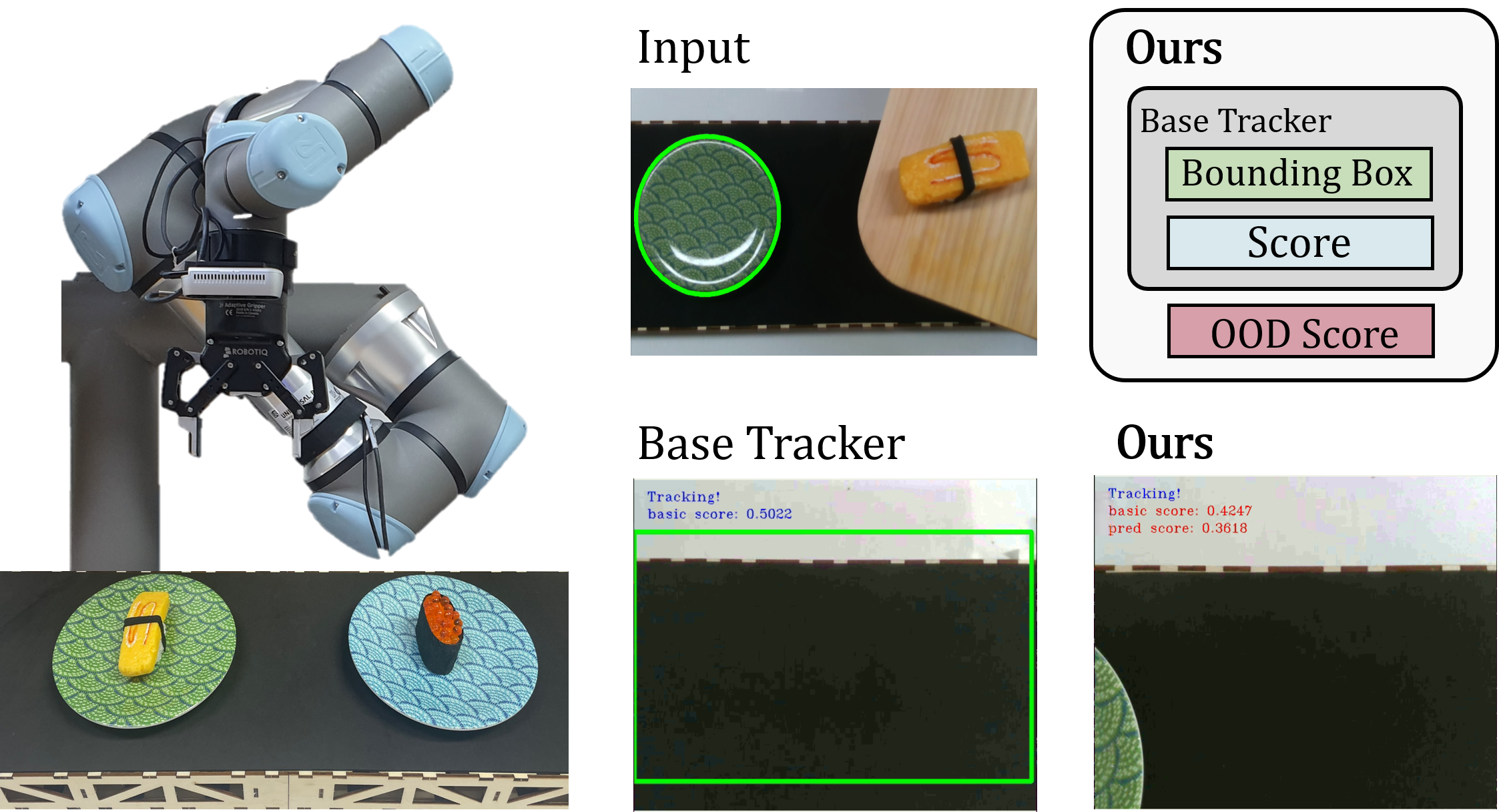}
\caption{\small The outline and the result of conveyor-belt sushi task. 
The exit prediction result of base tracker shows it does not recognize \emph{exit} (as it has green bbox) and has \textbf{0.5022} confidence on the result. However, the exit prediction of our tracker shows \textbf{0.4247} score function and \textbf{0.3618} OOD score function, each correctly predicts the exit.}
\label{sushi_task}
\vspace{-5mm}
\end{figure}

We evaluate our method in the conveyor-belt sushi task using an UR5e robot with a Robotiq 2F-85 gripper attached to it. The robot has an Intel Realsense D435i camera on its gripper and uses the camera to track the designated plate. 
The outline and the result of the task are described in Fig.~\ref{sushi_task}. First, the commander draws the bounding boxes on the plate. Then, the robot takes the sushi from the chef and it tracks the assigned dish. When the customer takes the dish, the robot with our tracker recognizes the \emph{exit} of the plate and output the low confidence score. However, the robot with the base tracker did not recognize the \emph{exit} and output the high confidence score, signaling the robot to track the background. 
The EXOT tracker robot waits for the target plate to reappear and successfully places the sushi on the plate. Meanwhile, the STARK tracker robot puts the sushi on the conveyor belt, not on the plate, and fails the task.

\section{CONCLUSIONS}
In this paper, we introduce the EXit-aware Object Tracker (EXOT) for safe robotic manipulation. In order to manipulate with human flexibly, EXOT tracks the moving objects and responds to the changing situation only with a robot hand camera. 
In the experiments, we confirm that EXOT can cope with unsafe situation by combining a single object tracker with OOD classification technique. Furthermore, we present the custom robot dataset, RMOT-223, which is suitable for training exit-aware trackers. EXOT outperforms the state-of-the-art single object tracker, STARK, especially in \emph{exit-awareness}. 
As a limitation, 
the threshold parameter for detecting OOD should be examined per dataset type. 
However, 
EXOT can learn and adapt to the situation gradually by unifying the online learning of changing threshold to our proposed method. 
Our method can be applied to any robotics domain that requires safe robot actions, as it is robot framework agnostic. 

%





%



\bibliographystyle{ieeetr}
\bibliography{ref}

\begin{thebibliography}{10}

\bibitem{tremblay2018corl:dope}
J.~Tremblay, T.~To, B.~Sundaralingam, Y.~Xiang, D.~Fox, and S.~Birchfield,
  ``Deep object pose estimation for semantic robotic grasping of household
  objects,'' in {\em Conference on Robot Learning (CoRL)}, 2018.

\bibitem{gou2021rgb}
M.~Gou, H.-S. Fang, Z.~Zhu, S.~Xu, C.~Wang, and C.~Lu, ``Rgb matters: Learning
  7-dof grasp poses on monocular rgbd images,'' in {\em 2021 IEEE International
  Conference on Robotics and Automation (ICRA)}, pp.~13459--13466, IEEE, 2021.

\bibitem{danielczuk2020exploratory}
M.~Danielczuk, A.~Balakrishna, D.~S. Brown, S.~Devgon, and K.~Goldberg,
  ``Exploratory grasping: Asymptotically optimal algorithms for grasping
  challenging polyhedral objects,'' {\em arXiv preprint arXiv:2011.05632},
  2020.

\bibitem{camacho2021reward}
A.~Camacho, J.~Varley, A.~Zeng, D.~Jain, A.~Iscen, and D.~Kalashnikov, ``Reward
  machines for vision-based robotic manipulation,'' in {\em 2021 IEEE
  International Conference on Robotics and Automation (ICRA)},
  pp.~14284--14290, IEEE, 2021.

\bibitem{driess2021learning}
D.~Driess, J.-S. Ha, R.~Tedrake, and M.~Toussaint, ``Learning geometric
  reasoning and control for long-horizon tasks from visual input,'' in {\em
  2021 IEEE International Conference on Robotics and Automation (ICRA)},
  pp.~14298--14305, IEEE, 2021.

\bibitem{robotnavi}
R.~V. Soans, Ranjith, A.~Hegde, C.~Singh, and A.~Kumar, ``Object tracking robot
  using adaptive color thresholding,'' in {\em 2017 2nd International
  Conference on Communication and Electronics Systems (ICCES)}, pp.~790--793,
  2017.

\bibitem{rad2017bb8}
M.~Rad and V.~Lepetit, ``Bb8: A scalable, accurate, robust to partial occlusion
  method for predicting the 3d poses of challenging objects without using
  depth,'' in {\em Proceedings of the IEEE international conference on computer
  vision}, pp.~3828--3836, 2017.

\bibitem{xiang2017posecnn}
Y.~Xiang, T.~Schmidt, V.~Narayanan, and D.~Fox, ``Posecnn: A convolutional
  neural network for 6d object pose estimation in cluttered scenes,'' {\em
  arXiv preprint arXiv:1711.00199}, 2017.

\bibitem{tekin2018real}
B.~Tekin, S.~N. Sinha, and P.~Fua, ``Real-time seamless single shot 6d object
  pose prediction,'' in {\em Proceedings of the IEEE conference on computer
  vision and pattern recognition}, pp.~292--301, 2018.

\bibitem{dunnhofer2021first}
M.~Dunnhofer, A.~Furnari, G.~M. Farinella, and C.~Micheloni, ``Is first person
  vision challenging for object tracking?,'' in {\em Proceedings of the
  IEEE/CVF International Conference on Computer Vision}, pp.~2698--2710, 2021.

\bibitem{staple}
L.~Bertinetto, J.~Valmadre, S.~Golodetz, O.~Miksik, and P.~H. Torr, ``Staple:
  Complementary learners for real-time tracking,'' in {\em Proceedings of the
  IEEE conference on computer vision and pattern recognition}, pp.~1401--1409,
  2016.

\bibitem{endToend}
N.~Carion, F.~Massa, G.~Synnaeve, N.~Usunier, A.~Kirillov, and S.~Zagoruyko,
  ``End-to-end object detection with transformers,'' in {\em European
  conference on computer vision}, pp.~213--229, Springer, 2020.

\bibitem{Fan_2019_CVPR}
H.~Fan, L.~Lin, F.~Yang, P.~Chu, G.~Deng, S.~Yu, H.~Bai, Y.~Xu, C.~Liao, and
  H.~Ling, ``Lasot: A high-quality benchmark for large-scale single object
  tracking,'' in {\em Proceedings of the IEEE/CVF Conference on Computer Vision
  and Pattern Recognition (CVPR)}, June 2019.

\bibitem{kristan2020eighth}
M.~Kristan, A.~Leonardis, J.~Matas, M.~Felsberg, R.~Pflugfelder, J.-K.
  K{\"a}m{\"a}r{\"a}inen, M.~Danelljan, L.~{\v{C}}. Zajc,
  A.~Luke{\v{z}}i{\v{c}}, O.~Drbohlav, {\em et~al.}, ``The eighth visual object
  tracking vot2020 challenge results,'' in {\em European Conference on Computer
  Vision}, pp.~547--601, Springer, 2020.

\bibitem{huang2019got}
L.~Huang, X.~Zhao, and K.~Huang, ``Got-10k: A large high-diversity benchmark
  for generic object tracking in the wild,'' {\em IEEE Transactions on Pattern
  Analysis and Machine Intelligence}, vol.~43, no.~5, pp.~1562--1577, 2019.

\bibitem{kristan2021ninth}
M.~Kristan, J.~Matas, A.~Leonardis, M.~Felsberg, R.~Pflugfelder, J.-K.
  K{\"a}m{\"a}r{\"a}inen, H.~J. Chang, M.~Danelljan, L.~Cehovin,
  A.~Luke{\v{z}}i{\v{c}}, {\em et~al.}, ``The ninth visual object tracking
  vot2021 challenge results,'' in {\em Proceedings of the IEEE/CVF
  International Conference on Computer Vision}, pp.~2711--2738, 2021.

\bibitem{mixformer}
Y.~Cui, C.~Jiang, L.~Wang, and G.~Wu, ``Mixformer: End-to-end tracking with
  iterative mixed attention,'' 2022.

\bibitem{stark}
B.~Yan, H.~Peng, J.~Fu, D.~Wang, and H.~Lu, ``Learning spatio-temporal
  transformer for visual tracking,'' in {\em Proceedings of the IEEE/CVF
  International Conference on Computer Vision}, pp.~10448--10457, 2021.

\bibitem{lin2021swintrack}
L.~Lin, H.~Fan, Y.~Xu, and H.~Ling, ``Swintrack: A simple and strong baseline
  for transformer tracking,'' {\em arXiv preprint arXiv:2112.00995}, 2021.

\bibitem{Damen2018EPICKITCHENS}
D.~Damen, H.~Doughty, G.~M. Farinella, S.~Fidler, A.~Furnari, E.~Kazakos,
  D.~Moltisanti, J.~Munro, T.~Perrett, W.~Price, and M.~Wray, ``Scaling
  egocentric vision: The epic-kitchens dataset,'' in {\em European Conference
  on Computer Vision (ECCV)}, 2018.

\bibitem{wang2020symbiotic}
X.~Wang, Y.~Wu, L.~Zhu, and Y.~Yang, ``Symbiotic attention with privileged
  information for egocentric action recognition,'' in {\em Proceedings of the
  AAAI Conference on Artificial Intelligence}, vol.~34, pp.~12249--12256, 2020.

\bibitem{liu2020forecasting}
M.~Liu, S.~Tang, Y.~Li, and J.~M. Rehg, ``Forecasting human-object interaction:
  joint prediction of motor attention and actions in first person video,'' in
  {\em European Conference on Computer Vision}, pp.~704--721, Springer, 2020.

\bibitem{sun2019deep}
S.~Sun, N.~Akhtar, H.~Song, A.~Mian, and M.~Shah, ``Deep affinity network for
  multiple object tracking,'' {\em IEEE transactions on pattern analysis and
  machine intelligence}, vol.~43, no.~1, pp.~104--119, 2019.

\bibitem{luo2021multiple}
W.~Luo, J.~Xing, A.~Milan, X.~Zhang, W.~Liu, and T.-K. Kim, ``Multiple object
  tracking: A literature review,'' {\em Artificial Intelligence}, vol.~293,
  p.~103448, 2021.

\bibitem{emami2018machine}
P.~Emami, P.~M. Pardalos, L.~Elefteriadou, and S.~Ranka, ``Machine learning
  methods for solving assignment problems in multi-target tracking,'' 2018.

\bibitem{attention}
A.~Vaswani, N.~Shazeer, N.~Parmar, J.~Uszkoreit, L.~Jones, A.~N. Gomez,
  {\L}.~Kaiser, and I.~Polosukhin, ``Attention is all you need,'' {\em Advances
  in neural information processing systems}, vol.~30, 2017.

\bibitem{generalized_ODIN}
Y.-C. Hsu, Y.~Shen, H.~Jin, and Z.~Kira, ``Generalized odin: Detecting
  out-of-distribution image without learning from out-of-distribution data,''
  2020.

\bibitem{TREK150}
M.~Dunnhofer, A.~Furnari, G.~M. Farinella, and C.~Micheloni, ``Is first person
  vision challenging for object tracking?,'' in {\em Proceedings of the
  IEEE/CVF International Conference on Computer Vision (ICCV) Workshops}, Oct
  2021.

\bibitem{milan2016mot16}
A.~Milan, L.~Leal-Taix{\'e}, I.~Reid, S.~Roth, and K.~Schindler, ``Mot16: A
  benchmark for multi-object tracking,'' {\em arXiv preprint arXiv:1603.00831},
  2016.

\bibitem{bai2021gmot}
H.~Bai, W.~Cheng, P.~Chu, J.~Liu, K.~Zhang, and H.~Ling, ``Gmot-40: A benchmark
  for generic multiple object tracking,'' in {\em Proceedings of the IEEE/CVF
  Conference on Computer Vision and Pattern Recognition}, pp.~6719--6728, 2021.

\bibitem{oodsurvey}
J.~Yang, K.~Zhou, Y.~Li, and Z.~Liu, ``Generalized out-of-distribution
  detection: A survey,'' 2021.

\bibitem{detr}
N.~Carion, F.~Massa, G.~Synnaeve, N.~Usunier, A.~Kirillov, and S.~Zagoruyko,
  ``End-to-end object detection with transformers,'' 2020.

\bibitem{revisitDetect}
G.~Song, Y.~Liu, and X.~Wang, ``Revisiting the sibling head in object
  detector,'' 2020.

\bibitem{revisitRCNN}
B.~Cheng, Y.~Wei, H.~Shi, R.~Feris, J.~Xiong, and T.~Huang, ``Revisiting rcnn:
  On awakening the classification power of faster rcnn,'' 2018.

\end{thebibliography}

\end{document}